\documentclass[conference]{IEEEtran}
\IEEEoverridecommandlockouts
\usepackage{cite}
\usepackage{amsmath,amssymb,amsfonts}
\usepackage{algorithmic}
\usepackage{graphicx}
\usepackage{textcomp}
\usepackage{xcolor}
\usepackage{tikz}
\usetikzlibrary{shapes.geometric, arrows}
\def\BibTeX{{\rm B\kern-.05em{\sc i\kern-.025em b}\kern-.08em
    T\kern-.1667em\lower.7ex\hbox{E}\kern-.125emX}}

\usepackage{float}
\usepackage{caption}
\begin{document}

\title{Adaptive Object Detection for Indoor Navigation Assistance: A Performance Evaluation of Real-Time Algorithms\\
% {\footnotesize \textsuperscript{*}Note: Sub-titles are not captured in Xplore and should not be used}
% \thanks{Identify applicable funding agency here. If none, delete this.}
}

\author
{\IEEEauthorblockN{Abhinav Pratap}
\IEEEauthorblockA{\textit{Department of Computer Science} \\
{\textit{and Engineering, ASET}} \\
\textit{Amity University, Noida, India} \\
TheAPratap@gmail.com}
\and
\IEEEauthorblockN{Sushant Kumar}
\IEEEauthorblockA{\textit{Department of Computer Science} \\
{\textit{and Engineering, ASET}} \\
\textit{Amity University, Noida, India} \\
sushantmishra02102002@gmail.com}
\and
\IEEEauthorblockN{Suchinton Chakravarty}
\IEEEauthorblockA{\textit{Department of Computer Science} \\
{\textit{and Engineering, ASET}} \\
\textit{Amity University, Noida, India} \\
suchinton.2001@gmail.com}
}

\maketitle

\begin{abstract}
This study addresses the critical need for accurate and efficient object detection in assistive technologies for visually impaired individuals. We systematically evaluate the performance of four prominent real-time object detection algorithms—YOLO, SSD, Faster R-CNN, and Mask R-CNN—within the context of indoor navigation assistance. Our analysis, conducted on the Indoor Objects Detection dataset, focuses on key parameters including detection accuracy, processing speed, and adaptability to the unique challenges of indoor environments. The findings highlight the crucial trade-off between precision and processing efficiency, providing valuable insights for selecting optimal algorithms for real-time assistive navigation systems. This research contributes to a deeper understanding of adaptive machine learning applications that can significantly improve indoor navigation solutions for the visually impaired, promoting inclusivity and accessibility.
\end{abstract}


\section{Introduction}
In today’s technology-driven society, there is an increasing emphasis on enhancing accessibility for visually impaired individuals. Indoor navigation poses unique challenges due to confined spaces, dynamic obstacles, and the need for precise and real-time object detection. While various object detection algorithms—such as YOLO [2, 8], SSD [2, 3], Faster R-CNN [7], and Mask R-CNN [7]—have been extensively studied in general computer vision contexts, their specific application to assistive indoor navigation remains underexplored. 

Recent studies have demonstrated the potential of object detection models in improving accessibility technologies. For example, Tinier-YOLO [11] and MobileNet-SSD [3] have shown promise in balancing accuracy and computational efficiency, making them viable candidates for resource-constrained environments. Similarly, research on assistive technologies for visually impaired individuals [13, 14] highlights the need for precise object detection in indoor settings to ensure safe and efficient navigation. However, these studies often lack a comprehensive evaluation of algorithm performance under the unique conditions of indoor environments, such as cluttered spaces and varying lighting conditions. 

This research aims to bridge this gap by systematically evaluating real-time object detection algorithms—YOLO, SSD, Faster R-CNN, and Mask R-CNN—on key parameters relevant to indoor navigation for visually impaired users. By focusing on the trade-offs between speed, accuracy, and adaptability, this study seeks to provide actionable insights for developing inclusive navigation systems that cater to the specific needs of this demographic.

\section{Dataset}
The Indoor Objects Detection dataset was created with the aim of assisting persons with visual impairments in their day-to-day life. This dataset supports the ‘Object Detection for Blind People’ project under the AI Builders 2022 initiative. The project's ultimate goal is the detection of indoor objects, emphasizing its relevance and applicability across various domains.

The choice to utilize the Indoor Objects Detection dataset for testing our models in object detection stems from its intrinsic relevance to our research focus on enhancing navigation for individuals with visual impairments. This dataset specifically targets indoor environments, aligning seamlessly with the challenges faced by visually impaired persons in navigating confined spaces. By encompassing 7331 labeled objects across 10 different indoor classes, including doors, cabinets, and furniture, the dataset provides a distinct, and broad set of scenarios for testing the effectiveness of our models. The inclusion of bounding box annotations further facilitates precise object detection, crucial for developing navigation assistance systems [13, 14]. Leveraging this dataset allows us to tailor our models to the unique demands of indoor settings, fostering the creation of more accurate and adaptive solutions for real-time navigation assistance. The average areas for classes are shown in Fig. 1. The Co-Occurrence Matrix of class objects in the dataset makes this dataset more useful for our application.

\begin{figure}[H]
\centerline{\includegraphics[scale=0.50]{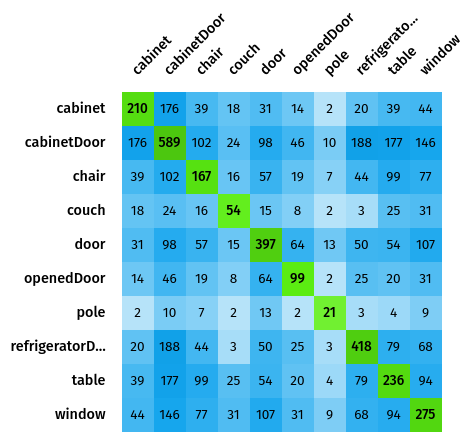}}
\caption{Co-Occurrence Matrix of class object in Dataset}
\label{dataset_matrix}
\end{figure}

\section{Related Work}
Object detection is a cornerstone of computer vision, and its applications have evolved significantly in recent years. YOLO, introduced as a real-time object detection model, revolutionized the field by achieving high detection speeds without compromising accuracy [2, 8]. Subsequent iterations, such as Tinier-YOLO, have optimized the model for constrained environments [11]. Similarly, SSD leverages predefined bounding boxes to balance speed and accuracy, making it a popular choice for dynamic settings [3].

Faster R-CNN and Mask R-CNN, while more computationally intensive, offer higher accuracy levels and detailed object segmentation, respectively. Studies like [6, 7] underscore their suitability for tasks requiring precision, but their applicability to real-time scenarios remains limited by their processing demands. Recent work on indoor navigation systems for visually impaired individuals [13, 14] highlights the critical role of object detection in addressing challenges such as identifying small objects and navigating crowded spaces. However, these studies often focus on a single algorithm or lack a comparative analysis across multiple models. 

Despite these advancements, existing research rarely explores the integration of these algorithms in assistive technologies tailored to indoor navigation. This gap is particularly pronounced in studies addressing real-time requirements and confined environments. Our work builds on these findings by conducting a comparative analysis of four prominent object detection models, emphasizing their performance in indoor scenarios and their implications for assistive navigation systems.  

The reviewed literature underscores significant progress in object detection and its applications. However, the unique challenges of indoor navigation—such as cluttered environments, dynamic obstacles, and the need for real-time responses—necessitate further investigation. By evaluating YOLO, SSD, Faster R-CNN, and Mask R-CNN under these specific conditions, this research addresses the limitations identified in prior studies and provides a robust framework for selecting algorithms that balance accuracy, speed, and adaptability.

\section{Algorithms}
\subsection{YOLO (You Only Look Once)}
YOLO, standing for You Only Look Once, revolutionizes real-time object detection by employing a grid-based approach. The input image is split into an SxS grid, where every grid cell is tasked with directly forecasting bounding boxes and class probabilities. This innovative strategy eliminates the need for a multi-step process, enabling YOLO to achieve high speed and efficiency. The algorithm's single-pass detection system contributes to its suitability for real-time applications, making it a favorite choice in various domains [2, 8].

\begin{figure}[H]
\centerline{\includegraphics[scale=0.37]{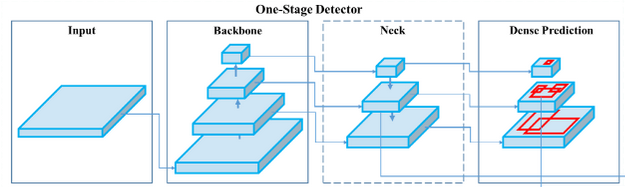}}
\caption{Single stage YOLO architecture}
\label{yolo_arc}
\end{figure}

A comprehensive working model of YOLO for a single frame is explained in Table~\ref{tab:yolo_working}:
\begin{table}[H]
\caption{Working of YOLO}
\label{tab:yolo_working}
\centering
\begin{tabular}{p{2.2cm} | p{4.8cm}}
\hline
\multicolumn{1}{c|}{\textbf{Steps}} &
\multicolumn{1}{c}{\textbf{Description}} \\
\hline
Input Image & Original image to be processed. \\
Grid Division & Image is divided into an S$\times$S grid. Each cell is responsible for object detection. \\
Bounding Box Prediction & Each grid cell predicts bounding boxes, class probabilities, and confidence scores. \\
Non Maximum Suppression & Redundant detections are filtered out using NMS based on confidence scores. \\
Output & Final bounding boxes, class probabilities, and confidence scores. \\
\hline
\end{tabular}
\end{table}

While the explanation used a single input image for simplicity, YOLO is commonly used in a batch processing mode. Instead of processing images one by one, YOLO can efficiently process multiple frames simultaneously. This parallel processing allows for real-time detection in video streams by handling multiple frames concurrently [2, 8].

\subsection{SSD (Single Shot Multibox Detector)}
SSD, or Single Shot MultiBox Detector, is an object detection technique optimized for a trade-off between accuracy and speed. It operates by employing a predefined set of boxes with various aspect ratios at each cell of the feature map. Simultaneously, it predicts both bounding box offsets and class scores.

\noindent Let's see the working of SSD in detail:
\begin{itemize}
    \item Feature Extraction: The input image is fed into a base Convolutional Neural Network (CNN), often based on architectures like VGG16. This CNN extracts feature maps that represent the hierarchical properties of the input image [3, 7].
    \item Default Box Assignment: At each position in the feature maps, default boxes of different aspect ratios are assigned. The SSD model then makes predictions for each of these default boxes, allowing it to handle objects of different shapes and sizes [3, 7].
    \item Prediction: For each default box, the SSD model simultaneously predicts the bounding box offsets and class scores. The bounding box offsets adjust the dimensions of the default box to better fit the actual location and size of the object. The class scores represent the confidence in the appearance of different object classes within the adjusted box [3, 7].
    \item Post-processing: After these predictions, a confidence threshold is applied to filter out predictions with low confidence scores. Additionally, Non-Maximum Suppression (NMS) is employed to eliminate redundant bounding box predictions, retaining only the most accurate ones [3, 7].
\end{itemize}

SSD's balanced accuracy and speed make it appropriate for real-time navigation systems. Its ability to handle various object scales in a single pass is advantageous, especially in dynamic environments. However, the potential decrease in accuracy for small objects might be a consideration depending on the specific requirements of a navigation application. The slight increase in computational intensity compared to YOLO should be weighed against the benefits of its versatile and efficient single-shot approach [3, 7]. 

\subsection{Faster R-CNN (Region-based Convolutional Neural Network)}
Faster R-CNN is a two-stage object detection framework which incorporates a Region Proposal Network (RPN) for generating region proposals and a Convolutional Neural Network (CNN) for feature extraction and classification.

\noindent Both processes are described below:
\begin{itemize}
    \item Region Proposal Network (RPN): Region Proposal Network (RPN) serves the function of suggesting potential object regions within an image. It accomplishes this by systematically moving a small window (anchor) across the feature map and predicting the presence or absence of an object. Concurrently, it adjusts the anchor boxes by refining them according to predicted offsets [6, 7].
    \item Feature Extraction and Classification (Faster R-CNN): Following the generation of region proposals, a Convolutional Neural Network (CNN) is utilized to extract features from each proposed region. These features are subsequently employed for both classification and bounding box regression tasks. Notably, the CNN shares convolutional layers with the Region Proposal Network (RPN), thereby improving computational efficiency [6, 7].
\end{itemize}

While Faster R-CNN excels in accuracy and precision, its slower processing speed and complexity may pose challenges in real-time navigation systems. For applications requiring instant responses and rapid object detection, such as autonomous navigation or assistive technologies for the visually impaired, speed limitations may be a critical factor [6, 7]. However, in controlled environments where high accuracy is prioritized over speed, Faster R-CNN could still find applications in navigation systems. The choice between Faster R-CNN and other algorithms depends on the explicit requirements and constraints of the navigation scenario [6, 7].

\subsection{Mask R-CNN}
Mask R-CNN builds on the Faster R-CNN framework by adding a new branch to predict segmentation masks. This allows Mask R-CNN to not just detect objects but also precisely outline and segment each distinct instance found within the detected regions.

\noindent Let's look at how Mask R-CNN works in more detail:
\begin{itemize}
    \item Region Proposal: Similar to Faster R-CNN, Mask R-CNN starts by generating region proposals using the Region Proposal Network (RPN) [6, 7].
    \item Bounding Box Prediction: The bounding box prediction branch operates the same way as in Faster R-CNN, refining the proposed bounding boxes and classifying the objects [6, 7].
    \item Segmentation Mask Prediction: The additional segmentation mask branch runs in parallel to predict pixel-level masks for each proposed bounding box. These masks provide detailed instance-level segmentation of the detected objects [6, 7].
    \item Post-processing: After making the predictions, post-processing steps like Non-Maximum Suppression (NMS) are applied to filter and refine the final set of bounding boxes and segmentation masks [6, 7].
\end{itemize}

Mask R-CNN's strength lies in tasks that demand detailed instance segmentation. While its accuracy in delineating objects is beneficial, the computational demands may be a consideration for real-time navigation systems, particularly in resource-constrained environments. Depending on the specific requirements of a navigation application, the trade-off between accuracy and computational efficiency should be carefully assessed to determine the suitability of Mask R-CNN [6, 7]. 

\section{Analysis}
Our analysis employs the Indoor Objects Detection dataset, focusing on evaluating each model’s performance across key parameters. All four models were analyzed for our use case. We used 107 indoor images from the test dataset and calculated the speed, total time, and average time of prediction for each model. We chose this dataset because it is well-suited for indoor navigation, and we wanted our model to be limited to this dataset as it is specifically used for navigation in indoor spaces [1, 4]. The test results are informative and can be used while building a navigation system for visually impaired individuals [1, 4].

\noindent Two primary metrics were utilized to gauge the models' detection performance:
\begin{enumerate}
    \item Intersection over Union (IoU): Measures overlap between predicted and ground-truth bounding boxes. IoU is computed as:
    \[
        \textit{IoU} = \frac{\textit{Area of Overlap}}{\textit{Area of Union}}
    \]
    Used for evaluating localization accuracy in object detection [3, 7].

    \item Mean Average Precision (mAP): Aggregated measure of precision across IoU thresholds. Average Precision (AP) at a specific IoU threshold:
    \[
        \textit{AP} = \frac{\textit{TP}}{\textit{TP} + \textit{FP}}
    \]
    mAP is the mean of AP values across recall levels and is a standard metric in object detection [3, 7].
\end{enumerate}

\begin{table}[H]
\caption{Validation Accuracy}
\label{tab:validation_accuracy}
\centering
\begin{tabular}{p{2.1cm}|c|c|c}
\hline
\multicolumn{1}{c|}{\textbf{Model}} &
\textbf{Dataset} &
\multicolumn{2}{c}{\textbf{Metrics}} \\
\cline{3-4}
 &  & \textit{Avg. IoU} & \textit{mAP} \\
\hline
YOLOv5 & COCO128 & – & 0.5 \\
\hline
SSD & & 0.5:0.95 & 0.195 \\
\cline{1-1}\cline{3-4}
Faster R-CNN & {2017 COCO} & 0.5:0.95 & 0.353 \\
\cline{1-1}\cline{3-4}
Mask R-CNN &  & 0.5:0.95 & 0.327 \\
\hline
\end{tabular}
\end{table}

\begin{table}[H]
\caption{Comparison of Models}
\label{tab:comparison_models}
\centering
\resizebox{7cm}{!}{%
\begin{tabular}{p{2.0cm} | c | c | c }
\hline
\multicolumn{1}{c|}{\textbf{Model}} &
\multicolumn{1}{c|}{\textbf{Accuracy}} &
\multicolumn{1}{c|}{\textbf{Total Time}} &
\multicolumn{1}{c}{\textbf{Avg. Time}} \\
\hline
YOLOv5 & 90.1\% & 38.25 sec & 0.357 sec \\
SSD & 79.8\% & 77.27 sec & 0.72 sec \\
Faster R-CNN & 69\% & 703.73 sec & 6.57 sec \\
Mask R-CNN & 61.06\% & 790.80 sec & 7.60 sec \\
\hline
\end{tabular}
}
\end{table}

The conducted analysis on the four object detection models—YOLOv5, SSD, Faster R-CNN, and Mask R-CNN—provides valuable insights into their performance on the selected Indoor Objects Detection dataset. The focus of the evaluation was on accuracy, total processing time, and average prediction time, all of which are critical factors for the development of a navigation system designed specifically for visually impaired people within indoor environments [1, 4, 7].

\subsection{Accuracy}
\begin{itemize}
    \item YOLOv5 (You Only Look Once): Demonstrates the highest accuracy among the models, reaching 90.1\%. This signifies YOLOv5's effectiveness in accurately detecting indoor objects within the specified dataset.
    \item SSD (Single Shot Multibox Detector): Achieves a commendable accuracy of 79.8\%, showcasing its capability to balance accuracy and speed for indoor object detection.
    \item Faster R-CNN (Region-based Convolutional Neural Network): Displays an accuracy of 69\%, representing a trade-off between accuracy and speed compared to YOLOv5 and SSD.
    \item Mask R-CNN: Exhibits a lower accuracy of 61.06\%, likely influenced by the additional complexity introduced by its instance segmentation capabilities.
\end{itemize}

\subsection{Average Prediction Time}
\begin{itemize}
    \item YOLOv5: Boasts the lowest average time at 0.357 seconds, reinforcing its suitability for real-time applications.
    \item SSD: Presents an average time of 0.72 seconds, indicating a reasonable balance between speed and accuracy.
    \item Faster R-CNN: Exhibits an average time of 6.57 seconds, showcasing a comparable performance to SSD.
    \item Mask R-CNN: Records an average time of 7.60 seconds, reinforcing the computational demands associated with its instance segmentation capability.
\end{itemize}

YOLOv5 emerged as the most balanced model, achieving 90.1\% accuracy with an average prediction time of 0.357 seconds, making it ideal for real-time applications. SSD, with an accuracy of 79.8\% and a prediction time of 0.72 seconds, also demonstrated a reasonable balance. In contrast, Faster R-CNN and Mask R-CNN, while achieving higher precision, exhibited slower processing times, limiting their utility in real-time navigation. 

These findings align with earlier studies emphasizing the importance of speed in assistive technologies [11, 13]. However, our study extends this understanding by demonstrating that models like YOLOv5 can effectively address the specific challenges of indoor environments, such as dynamic obstacles and varying object scales. This highlights the necessity of tailoring algorithm selection to the unique demands of the application. 

By integrating insights from prior research with empirical evaluation, this study bridges the gap between theoretical advancements in object detection and practical applications in assistive navigation. Our work provides a holistic framework for evaluating and implementing object detection models in real-world scenarios.

\section{Conclusions}
This research contributes to the expanding field of assistive technologies by providing a detailed evaluation of object detection algorithms for indoor navigation. Among the models tested, YOLOv5 emerged as the most suitable for real-time applications, offering an optimal balance of speed and accuracy with 90.1\% accuracy and an average prediction time of 0.357 seconds. SSD, while slightly less accurate (79.8\%), also demonstrated strong performance, making it another viable choice for real-time navigation systems.

However, models like Faster R-CNN and Mask R-CNN, despite offering higher accuracy, presented significant computational challenges. The trade-off between precision and processing time underscores the importance of balancing these factors when developing systems for visually impaired users, where real-time efficiency is paramount.

This research not only adds valuable insights into the comparative performance of these models but also lays the foundation for future advancements in assistive technology. Future research could focus on integrating these models with advanced sensor systems or adapting them for outdoor navigation scenarios, further enhancing their practical applications and impact.

\section*{Acknowledgment}
We extend our deepest gratitude to all the researchers, healthcare workers, and practitioners who have significantly contributed to advancements in assistive technologies and public health. We are also grateful to the developers and maintainers of the Indoor Objects Detection dataset [13, 14], whose efforts have made this study possible. Their invaluable work in creating robust datasets and frameworks has enabled meaningful strides in indoor navigation assistance for visually impaired individuals.

\vspace{12pt}

\end{document}